\documentclass{article}
\usepackage{stfloats}
\usepackage{spconf,amsmath,graphicx,hyperref}
\usepackage{amsfonts}
\usepackage{booktabs}
\usepackage[table]{xcolor}
\usepackage{multirow, makecell, threeparttable, tabularx}
\usepackage{siunitx}
\newcolumntype{Y}{S[table-format=2.2(2)]} % 例如 88.27 ± 0.08

\definecolor{HeaderGray}{HTML}{D0CECE}
\definecolor{Stripe}{HTML}{E7E6E6}
\definecolor{OursBg}{HTML}{FFF2CC}

\newcommand{\inc}[1]{\textcolor{red}{\(\uparrow\)#1}}
\newcommand{\dec}[1]{\textcolor{green!60!black}{\(\downarrow\)#1}}
\newcommand{\best}[1]{\textbf{#1}}

\newcolumntype{L}{!{\vrule width \heavyrulewidth}}
\newcolumntype{G}{@{}p{2pt}@{}}

\everymath{\small}
\everydisplay{\small}

\title{FedDBP: Enhancing Federated Prototype Learning with Dual-Branch Features and Personalized Global Fusion}

\name{Ningzhi Gao, Siquan Huang, Leyu Shi, Ying Gao$^{\ast}$ \thanks{*Corresponding author. This work is supported by the National Natural Science Foundation of China (Grant No. 62476095), and Guangdong Basic and Applied Basic Research Foundation (Grant No. 2025A1515011525).}}
\address{South China University of Technology, Guangzhou, China}
\begin{document}
\ninept
\maketitle
\begin{abstract}
% The abstract should appear at the top of the left-hand column of text, about
% 0.5 inch (12 mm) below the title area and no more than 3.125 inches (80 mm) in
% length.  Leave a 0.5 inch (12 mm) space between the end of the abstract and the
% beginning of the main text.  The abstract should contain about 100 to 150
% words, and should be identical to the abstract text submitted electronically
% along with the paper cover sheet.  All manuscripts must be in English, printed
% in black ink.
Federated prototype learning (FPL), as a solution to heterogeneous federated learning (HFL), effectively alleviates the challenges of data and model heterogeneity. 
% FPL leverages prototypes as carriers, enabling clients to share knowledge with one another. 
However, existing FPL methods fail to balance the fidelity and discriminability of the feature, and are limited by a single global prototype. 
In this paper, we propose \textbf{FedDBP}, a novel FPL method to address the above issues. On the client-side, we design a {\bf D}ual-{\bf B}ranch feature projector that employs L2 alignment and contrastive learning simultaneously, thereby ensuring both the fidelity and discriminability of local features. On the server-side, we introduce a {\bf P}ersonalized global prototype fusion approach that leverages Fisher information to identify the important channels of local prototypes. Extensive experiments demonstrate the superiority of FedDBP over ten existing advanced methods.
\end{abstract}
\begin{keywords}
heterogeneous federated learning, prototype learning, personalization
\end{keywords}

\section{Introduction}
\label{sec:intro}

% These guidelines include complete descriptions of the fonts, spacing, and
% related information for producing your proceedings manuscripts. Please follow
% them and if you have any questions, direct them to Conference Management
% Services, Inc.: Phone +1-979-846-6800 or email
% to \\\texttt{papers@2026.ieeeicassp.org}.

Federated learning (FL) \cite{ref1} allows multiple clients to train a global model without sharing data. Conventional FL frameworks are usually based on idealized assumptions \cite{ref1}, e.g., the same data distribution or model architecture across clients. Unfortunately, real-world FL scenarios usually encounter challenges arising from data or model heterogeneity \cite{zhao2018federated,li2020federated,diao2020heterofl,huang2025fedid}. Data and model heterogeneity causes the global model to suffer from significant performance degradation or invalidation \cite{zhao2018federated,ref22}. Hence, heterogeneous federated learning (HFL) has been proposed and widely applied \cite{ref3,ref6}.

Existing HFL methods are roughly categorized into knowledge distillation-based \cite{ref4,ref5}, model decomposition-based \cite{ref7,ref8}, generator-based \cite{ref9,ref13,ref21}, and feature alignment \cite{ref10,ref12}. 
% Each facilitates cross-client knowledge transfer but suffers respectively from dependence on auxiliary data quality, architectural similarity constraints, and high computational cost. 
Among these methods, FedProto \cite{ref12} pioneered federated prototype learning (FPL), which realizes knowledge transfer through prototype alignment. It supports fully heterogeneous models and significantly reduces the communication cost. Specifically, proposed FPL studies \cite{ref12,ref14,ref15,ref16,ref17} are divided into client-side and server-side. On the client-side, L2 alignment–based methods \cite{ref12,ref16} align the local features with global prototypes, which ensures the feature fidelity. Contrastive learning–based methods \cite{ref14,ref17} pull the positive pairs closer, enhancing the discriminability of features. Some methods like \cite{ref15} combine L2 alignment with contrastive learning to enlarge inter-class separability without deviating from the overall direction. On server-side, existing methods generate high-quality global prototypes through approaches such as clustering \cite{ref15,ref17} or learnable networks \cite{ref16}. Although existing methods have improved the performance of FPL, existing client-side methods overlook the balance between feature fidelity and discriminability, and server-side approaches are limited by a single global prototype.

For client-side approaches, the L2 alignment ensures the feature fidelity while neglecting the relative distances among features of different classes. Also, contrastive learning enhances the feature discriminability by focusing on the relative distances between representations, while leading to low-quality global prototypes due to deviation from semantic centers. Although \cite{ref15} employs both alignment strategies simultaneously, the conflict of two measures on the same feature extractor may cause confusion. In light of these analyses, we raise {\bf issue (i)} in client-side approaches: \textit{How to simultaneously ensure both the fidelity and the discriminability of local features?} 
On the server-side, existing FPL methods are limited by a single global prototype. The proposed research \cite{ref22} has shown that data from the same class exhibit different feature distributions across clients due to data heterogeneity. Single global prototype mixes the semantics of different clients, which results in the loss of client-specific local information. Such semantically ambiguous global prototypes mislead the training of the clients, thereby degrading model performance. Based on this finding, we raise {\bf issue (ii)}: \textit{How to overcome the limitations of a single global prototype?}

To address \textbf{issue (i)}, we design a novel dual-branch feature extraction including a shared and a decision branch. Specifically, the shared branch adopts an L2 alignment–based strategy, which outputs the local prototypes and then uploads them to the server, ensuring the fidelity. The decision branch employs contrastive learning with hard negative mining, which produces high discriminability features, thereby improving the accuracy of prototype-based inference. For \textbf{issue (ii)}, we propose to construct personalized global prototypes by preserving client-specific local information.
The Fisher information quantifies the sensitivity of the output distribution to minimal changes in the feature channels. Inspired by Fisher information \cite{ref23,ref24,ref25}, we propose a channel importance–aware approach. We assume the channels with higher Fisher information contain more local information and assign them higher importance scores.
Thus, we employ a Top-K selection to perform channel-wise prototype fusion, which makes clients preserve the local important information. Comprehensively, we summarize our main contributions below:

\begin{figure*}[!t]          % 星号=跨双栏，[!t]=尽量放在页面顶部
\setlength{\abovecaptionskip}{-5cm}
\setlength{\belowcaptionskip}{-5cm}
  \centering
  \includegraphics[width=0.98\textwidth]{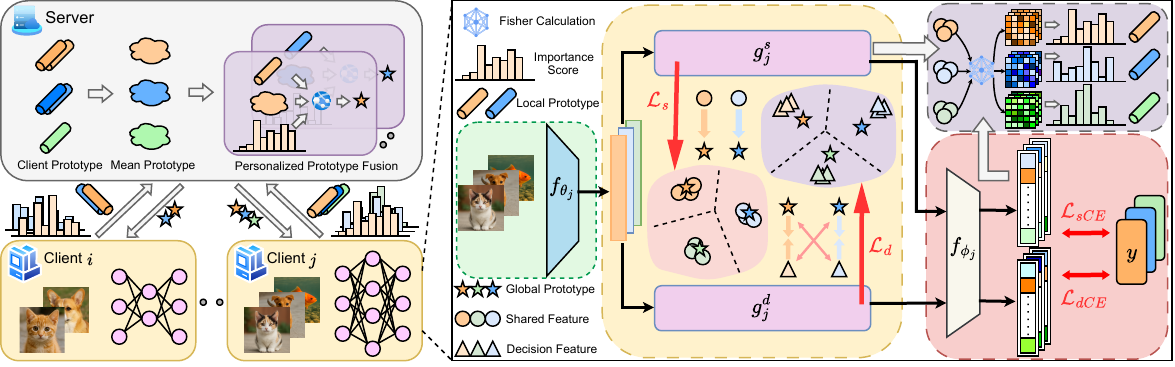}  % 用 \textwidth 占满双栏宽
  \caption{Overview of FedDBP. }
  \label{fig:placeholder}
\end{figure*}

\begin{itemize}
    \item We propose a novel FPL approach, namely FedDBP, including effective client training and adaptive server aggregation, which improves local feature quality and strengthens personalization of the global prototype, respectively.
    \item We design a client-side dual-branch feature projector, which integrates L2 alignment and contrastive learning simultaneously, ensuring feature fidelity and discriminability. 
    \item We introduce a Fisher information-based channel importance analysis, which assigns channel weights adaptively, achieving personalized global prototype fusion on the server.
    \item Extensive experiments under heterogeneous scenarios show that FedDBP achieves state-of-the-art performance on four datasets, compared with $10$ advanced HFL methods.
\end{itemize}

\section{PROBLEM FORMULATION}
\label{sec:problem}

\subsection{Problem Setting}
In the HFL scenario, we assume the presence of a central server and $K$ clients, denoted as $\mathcal{C} = \{1,2,\ldots,K\}$. Each client $k \in \mathcal{C}$ holds a local dataset $\mathcal{D}_k$, with the size $n_k = |\mathcal{D}_k|$, and the total number of samples is 
$N = \sum_{k=1}^{K} n_k$. The heterogeneity is reflected in data and model described as follows:
\begin{itemize}
    \item {\bf Data heterogeneity}: The local data of each client denoted as $P_k(x,y)$,  is non-independent and identically distributed (non-IID). For different clients, i.e., $k \neq j$, we have $P_k \neq P_j$.
    \item {\bf Model heterogeneity}: The model of client $k$ is parameterized by $\omega_k \in \mathbb{R}^{d_k}$. For different clients, i.e., $k \neq j$, both the dimensionality and the structure of model parameters are entirely different, meaning $d_k \neq d_j$.
\end{itemize}

\subsection{Optimization Objective} 
The optimization object of HFL is to learn a set of high-performing personalized models $\{\omega_1^*, \omega_2^*, \ldots, \omega_K^*\}$, such that the weighted sum of local loss functions across all clients is minimized. Formally, the objective can be expressed as:
\begin{equation}
\min_{\omega_1, \ldots, \omega_K} \mathcal{F}(\omega_1, \ldots, \omega_K) 
:= \sum_{k=1}^{K} \frac{n_k}{N} \mathcal{F}_k(\omega_k),
\end{equation}
where $\mathcal{F}_k(\omega_k)$ denotes the empirical loss of $k$-th client on its local dataset $\mathcal{D}_k$:
\begin{equation}
\mathcal{F}_k(\omega_k) := \frac{1}{n_k} \sum_{(x_i, y_i) \in \mathcal{D}_k} 
\ell_k(\omega_k; x_i, y_i),
\end{equation}
where $\ell_k$ denotes the local loss function of $k$-th client.

% \subsection{Federated Prototype Learning}
% \textbf{Tan et al.} proposed federated prototype learning, FedProto\cite{ref12}, which facilitates knowledge sharing among clients through prototype vectors. Each client model $\omega_k$ is viewed as consisting of two components: a feature extractor $\theta_k$ and a classification head $\phi_k$. 

% The class prototype $p_k^c$ is defined for class $c$ in client $k$ as the aggregation of features produced by the feature extractor for samples belonging to that class, serving as the encoded representation of the class:
% \begin{equation}
% p_k^c = \frac{1}{|\mathcal{D}_k^c|} \sum_{x_i \in \mathcal{D}_k^c} f_{\theta_k}(x_i).
% \end{equation}

\section{METHOD}
\label{sec:method}
We first give the overview of FedDBP in Fig. \ref{fig:placeholder}, which mainly consists of two components, \textbf{Dual-Branch Feature Projector} and \textbf{Personalized Global Prototype Fusion}. In the right side, data goes through shared branch and decision branch, respectively. Clients update their model by minimizing the cross-entropy loss $\mathcal{L}_\mathrm{sCE}$ and $\mathcal{L}_\mathrm{dCE}$,  L2 loss for shared branch $\mathcal{L}_s$, and the contrastive loss for decision branch $\mathcal{L}_d$. For personalized global prototype fusion, clients compute local prototypes and channel importance scores, and then uploads them to the server.
% The client data goes through the feature extractor, then the shared branch and the decision branch, respectively, and finally the classification head. The clients update their local model by minimizing the cross-entropy loss $\mathcal{L}_\mathrm{sCE}$ and $\mathcal{L}_\mathrm{dCE}$,  the Euclidean distance between shared feature and global prototype $\mathcal{L}_s$, and the contrastive loss between decision feature and global prototype $\mathcal{L}_d$. The client calculates the local prototypes and channel importance scores, and then uploads them to the server. The server calculates the mean prototype and performs the personalized global prototype fusion.

\subsection{Dual-Branch Feature Projector}
To balance feature fidelity and discriminability, we design a dual-branch feature projector on the client side. We introduce a shared branch $g^s$ and a decision branch $g^d$ between the feature extractor $f_{\theta}$ and the classification head $f_{\phi}$. Both branches optimize by cross-entropy loss $\mathcal{L}_\mathrm{sCE}$ and $\mathcal{L}_\mathrm{dCE}$, respectively.

Subsequently, we aim to leverage shared features from the shared branch $g^s$ to construct local prototypes of clients. Specifically, the local prototype $P_k^c$ is the representation of class $c$ of client $k$, which is computed by the average of all shared features $z^s$ belonging to that class. 
% These local prototypes are then uploaded to the server for aggregation. 
Since, these shared features should effectively represent the global prototypes 
$P = [p^1, p^2, \ldots, p^C]$, we design the loss functions $\mathcal{L}_s$. It aligns the shared branch features with their global prototypes in the semantic space by minimizing the Euclidean distance between them, denoted as:

\begin{equation}
\mathcal{L}_s = \frac{1}{B} \sum_{i=1}^{B} \left\| p^{y_i} - z_i^s \right\|_2^2,
\end{equation}

Moreover, we employ the decision feature $z^d$ from the decision branch $g^d$ to perform prototype-based 
% nearest neighbor 
inference. Since the decision branch should generate the features exhibiting clear inter-class discriminability, we apply contrastive learning to address this issue. We thereby design the following loss functions $\mathcal{L}_d$ for the decision branch, which enhances feature discriminability, denoted as:

\begin{equation}
\mathcal{L}_d = - \frac{1}{B} \sum_{i=1}^{B} 
\frac{\exp\left(-d_{i,y_i}/\tau\right)}
{\sum_{c=1}^{C} \exp\left(-d_{i,c}/\tau\right)},
\end{equation}
where $d_{i,c}$ denotes the Euclidean distance between the normalized decision branch feature $\hat{z}_i^d$ and the normalized global class prototype $\hat{p}^c$. 
The temperature parameter is $\tau$, and $C$ represents the total number of classes. 
% The loss $\mathcal{L}_d$ enhances feature discriminability through contrastive learning, encouraging decision branch features of different classes to be well separated.

Notably, assigning equal weights to easy and hard negative samples in $\mathcal{L}_d$ is ineffective. Easy negatives contribute little gradient, whereas hard negatives near class boundaries are more important for discriminating features. To address this, we introduce an adaptive hard negative mining strategy that dynamically selects challenging negatives to improve the effectiveness. Therefore, we incorporate a boundary penalty term $M_i$ into the denominator of the $\mathcal{L}_d$ loss, and redefine $\mathcal{L}_d$ as below:

\begin{equation}
\mathcal{L}_d = - \frac{1}{B} \sum_{i=1}^{B} 
\frac{\exp\left(-d_{i,y_i}/\tau\right)}
{\sum_{c=1}^{C} \exp\left(-d_{i,c}/\tau\right) + M_i},
\end{equation}
where
\begin{equation}
M_i = \sum_{c \neq y_i} \exp\left(-\frac{\max(0,\, m - d_{i,c})}{\tau}\right),
\end{equation}
where the adaptive boundary $m$ is defined as 
$m = \tfrac{\bar{d}^+ + \bar{d}^-}{2}$,
where $\bar{d}^+$ and $\bar{d}^-$ denote the average Euclidean distances among positive and negative samples in the batch, respectively. We compute the adaptive boundary $m$ based on the current batch statistics, ensuring that the boundary adapts to different data distributions and training processes.

Therefore, the overall objective of local training of FedDBP is defined as:
\begin{equation}
\mathcal{L} = \mathcal{L}_{\mathrm{sCE}} 
+ \lambda_1 \, \mathcal{L}_{\mathrm{dCE}} 
+ \lambda_2 \mathcal{L}_s 
+ \lambda_3 \mathcal{L}_d,
\end{equation}
where $\lambda_1$, $\lambda_2$ and $\lambda_3$ are hyperparameters that balance the contributions of the respective loss terms.

% The shared branch focuses on producing more representative shared class features $z^s$. After absorbing global knowledge and local training, the shared branch more accurately represents the absolute positions of classes in the low-dimensional feature space. Accordingly, we define the local prototype $P_k^c$ as the representative feature of class $c$ for client $k$, which is computed as the mean of all shared features belonging to class $c$. Once the local prototypes from all clients are collected, they are uploaded to the server. 
% \begin{equation}
% P_k^c = \frac{1}{N_k^c} \sum_{\substack{x_i \in \mathcal{D}_k, y_i = c}} z_i^s,
% \end{equation}
% where $N_k^c$ denotes the number of samples in client $k$ with class $c$, and $z_i^s$ represents the shared feature of the $i$-th sample in client $k$ belonging to class $c$.

% The decision branch focuses on producing discriminative decision features $z^d$. Thanks to the mathematical mechanism of the contrastive loss, features from different classes are well separated in the low-dimensional feature space and decision features provide clearer classification boundaries. Consequently, during prototype-based nearest neighbor inference, we compute the Euclidean distance between each decision feature $z^d$ and the global prototype of each class $P^c$, and assign the class corresponding to the nearest prototype as the predicted label $\hat{y}$.
% \begin{equation}
% \hat{y} = \arg\min_j \left\| z^d - P^j \right\|_2.
% \end{equation}

\subsection{Personalized Global Prototype Fusion}
Since heterogeneity makes single global prototype ineffective, we introduce a Fisher information–based approach, that generates client-specific global prototypes by estimating channel importance in the shared features $z^s$. The Fisher information quantifies the sensitivity of the output distribution to minimal changes in the feature channel. A larger Fisher value indicates that the channel has a significant impact on the client’s prediction, which should be preserved. We approximate importance scores as below:

% It is defined as:
% \begin{equation}
% I(\theta) = \mathbb{E} \left[ 
% \left( \frac{\partial \log p(x \mid \theta)}{\partial \theta} \right)
% \left( \frac{\partial \log p(x \mid \theta)}{\partial \theta} \right)^{\!T}
% \right].
% \end{equation}
\begin{equation}
s_{c} = \frac{1}{N_c} \sum_{i: y_i = c} 
\left( \frac{\partial \log p(y_i \mid x_i;\, \omega)}{\partial z_{i,j}} \right)^2,
\end{equation}
where $N_c$ denotes the number of samples belonging to class $c$, and $z_{i,j}$ is the $j$-th channel of the feature $z_i$.

% Specifically, after one round of local training, the client reuses its dataset $\mathcal{D}_i = \{(x_i, y_i)\}$ for forward propagation through the backbone, shared branch, and classifier to obtain logits $o_i = [o_{i,1},  \ldots, o_{i,C}]$. The log-probability of the true label is computed as $\log p(y_i \mid x_i) = o_{i,y_i} - \log \big( \sum_{k} \exp(o_{i,k}) \big)$. The gradient with respect to the shared feature $z_i^s$ is obtained by backpropagation as $\text{grad}_i = \frac{\partial \log p(y_i \mid x_i)}{\partial z_i^s} = [\text{grad}_{i,1}, \ldots, \text{grad}_{i,d_z}]$. The Fisher information for the $j$-th channel of class $c$ is then given by $F_{c,j}^i = (\text{grad}_{i,j})^2$. By averaging over all samples in class $c$, we obtain the class-level Fisher information $\bar{F}_{c,j} = \tfrac{1}{|\mathcal{D}_c|} \sum_{(x_i,y_i) \in \mathcal{D}_c} F_{c,j}^i$. Finally, the second-order Fisher information is rescaled to the first-order form $s_{c,j} = \sqrt{\bar{F}_{c,j} + \epsilon}$, which serves as the channel importance score.
 
 % A larger Fisher value indicates that the channel has a significant impact on the client’s prediction. Those channels carry more local characteristics, which should be preserved. 
 
During computing local prototypes, the client utilizes its dataset for forward propagation through the backbone, shared branch, and classifier to obtain logits. Then, we compute the log-probability of the true label. The gradient with respect to the shared feature is obtained by backpropagation. Based on these gradients, the Fisher information for each channel of class $c$ can be estimated. Finally, by averaging over all samples in the same class, we obtain the class-level Fisher information as channel importance scores $s \in \mathbb{R}^{C \times d_z}$.

% As mentioned in Section~3.1, once the local prototypes of each client $P_k = [P_k^1, P_k^2, \ldots, P_k^C]$ are collected, they are uploaded to the server. Similarly, the channel importance scores of each client are also transmitted to the server, to facilitate global prototype aggregation and the fusion of personalized global prototypes. 
In addition, we adopt client-wise averaging to compute the global prototype for each class: 
$p^c = \tfrac{1}{|M_c|} \sum_{k \in M_c} p_k^c$, 
where $|M_c|$ denotes the number of clients containing class $c$. For the fusion of personalized global prototypes, we leverage the channel importance scores of each client and employ a Top-$K$ selection strategy. Specifically, we retain the most important $K$ channels with more local information, while adopting the global prototype in the other channels, thereby achieving channel-wise personalized fusion. The personalized global prototypes replace the averaged global prototypes and are distributed to the participating clients.

\begin{equation}
T_c = \{ j \;|\; s_{c,j} \in \text{Top-}K\{s_{c,1}, \ldots, s_{c,d_z}\} \}.
\end{equation}
\begin{equation}
p_{k,\text{pers}}^{c,j} =
\begin{cases}
\eta \cdot p_k^{c,j} + (1 - \eta) \cdot p^{c,j}, & \text{if } j \in T_c \\
p^{c,j}, & \text{if } j \notin T_c
\end{cases}
\end{equation}
where $\eta$ is the fusion weight.

\section{EXPERIMENT}
\label{sec:exp}

\begin{table*}[!t]
\setlength{\abovecaptionskip}{0pt}
\setlength{\belowcaptionskip}{0pt}
\centering
\setlength\tabcolsep{6pt}
\renewcommand{\arraystretch}{1.15}
\caption{Comparison with state-of-the-art federated prototype learning and other HFL methods on four datasets. We show the average test accuracy(\%) across clients for each method. Best in bold and second with underline. The growth or reduction is compared to FedProto.}
\begin{threeparttable}
\label{tab:main}
\rowcolors{3}{Stripe}{white}

% 关键：方法列右对齐 + 第一/第二列之间竖线
\begin{tabular}{>{\raggedleft\arraybackslash}p{3.1cm}!{\vrule width \heavyrulewidth}*{4}{Y c}}
\Xhline{1.2pt} 
\rowcolor{HeaderGray}
\makecell{} &
\multicolumn{2}{c}{\textit{CIFAR-10}} &
\multicolumn{2}{c}{\textit{CIFAR-100}} &
\multicolumn{2}{c}{\textit{Flowers102}} &
\multicolumn{2}{c}{\textit{Tiny-ImageNet}} \\
\hline
\addlinespace[2pt]
\hline

FedProto {[}AAAI22{]} &
\multicolumn{2}{c}{\num{84.32 \pm 0.24}} &
\multicolumn{2}{c}{\num{39.01 \pm 0.17}} &
\multicolumn{2}{c}{\num{43.61 \pm 0.37}} &
\multicolumn{2}{c}{\num{19.33 \pm 0.08}} \\
\Xhline{1.2pt} 

\rowcolor{gray!12}
FedProc {[}FCGS23{]} &
\num{86.14 \pm 0.19} & \inc{1.82} &
\num{41.12 \pm 0.26} & \inc{2.11} &
\num{44.00 \pm 0.53} & \inc{0.39} &
\num{23.54 \pm 0.35} & \inc{4.21} \\

FPL {[}CVPR23{]} &
\num{86.08 \pm 0.14} & \inc{1.76} &
\num{40.15 \pm 0.12} & \inc{1.04} &
\num{40.33 \pm 0.20} & \dec{3.28} &
\num{20.87 \pm 0.15} & \inc{1.54} \\ 

\rowcolor{gray!12}
FedTGP {[}AAAI24{]} &
\underline{\num{88.04 \pm 0.21}} & \underline{\inc{3.72}} &
\num{41.77 \pm 0.27} & \inc{2.76} &
\underline{\num{51.74 \pm 0.31}} & \underline{\inc{8.13}} &
\num{24.72 \pm 0.11} & \inc{5.39} \\

FedLFP {[}TMC25{]} &
\num{85.76 \pm 0.09} & \inc{1.44} &
\num{39.80 \pm 0.18} & \inc{0.79} &
\num{47.38 \pm 0.23} & \inc{3.77} &
\num{23.37 \pm 0.21} & \inc{4.04} \\
\Xhline{1.2pt} 

\rowcolor{gray!12}
FD {[}arXiv18{]} &
\num{86.61 \pm 0.11} & \inc{2.29} &
\num{41.42 \pm 0.14} & \inc{2.41} &
\num{48.69 \pm 0.38} & \inc{5.08} &
\num{25.57 \pm 0.28} & \inc{6.24} \\

LG-FedAvg {[}NIPS19{]} &
\num{83.81 \pm 0.07} & \dec{0.51} &
\num{40.39 \pm 0.10} & \inc{1.38} &
\num{46.45 \pm 0.34} & \inc{2.84} &
\num{23.91 \pm 0.13} & \inc{4.58} \\

\rowcolor{gray!12}
FedGen {[}ICML21{]} &
\num{83.89 \pm 0.39} & \dec{0.43} &
\num{39.88 \pm 0.17} & \inc{0.87} &
\num{46.84 \pm 0.25} & \inc{3.23} &
\num{23.50 \pm 0.29} & \inc{4.17} \\

FedKD {[}NC22{]} &
\num{87.30 \pm 0.13} & \inc{2.98} &
\underline{\num{42.59 \pm 0.21}} & \underline{\inc{3.58}} &
\num{49.86 \pm 0.33} & \inc{6.25} &
\underline{\num{26.37 \pm 0.27}}    & \underline{\inc{7.04}} \\

\rowcolor{gray!12}
FedMRL {[}NIPS24{]} &
\num{86.49 \pm 0.25} & \inc{2.17} &
\num{41.94 \pm 0.33} & \inc{2.93} &
\num{47.58 \pm 0.18} & \inc{3.97} &
\num{23.94 \pm 0.35} & \inc{4.61} \\
\Xhline{1.2pt} 

\rowcolor{OursBg}
\textbf{FedDBP(ours)} &
\best{\num{88.27 \pm 0.08}} & \best{\inc{\textbf{3.95}}} &
\best{\num{45.59 \pm 0.12}} & \best{\inc{\textbf{6.58}}} &
\best{\num{52.99 \pm 0.24}} & \best{\inc{\textbf{9.40}}} &
\best{\num{27.14 \pm 0.19}} & \best{\inc{\textbf{7.86}}} \\
\Xhline{1.2pt} 
\end{tabular}
\end{threeparttable}
\end{table*}

\subsection{Experiment Settings}
{\bf Datasets and Models.} We conduct experiments on four image classification datasets: CIFAR-10   \cite{ref18}, CIFAR-100  \cite{ref18}, Flowers102 \cite{ref19}, and Tiny-ImageNet \cite{ref20}. To simulate data heterogeneity in HFL, we partition the data into non-IID subsets for each client using a Dirichlet distribution \cite{ref26} $\mathrm{Dir}(\alpha)$ with $\alpha=0.1$ by default. To simulate client model heterogeneity, we follow the settings in HtFLlib \cite{ref6}, using HtFE$_{8}^{\mathrm{img}}$.

{\bf Baselines.} Under scenarios of both data heterogeneity and model heterogeneity, we compare FedDBP with five state-of-the-art FPL methods (FedProto \cite{ref12}, FedProc \cite{ref14}, FPL \cite{ref15}, FedTGP \cite{ref16}, FedLFP \cite{ref17}) and five other HFL methods (LG-FedAvg \cite{ref7}, FD \cite{ref13}, FedGen \cite{ref9}, FedKD \cite{ref4}, FedMRL \cite{ref5}). 

{\bf Implementation Details.} All experiments are conducted on a server equipped with four NVIDIA GeForce RTX 2080 Ti GPUs and an Intel(R) Core(TM) i9-10920X CPU.
We use $20$ clients with a client participation rate of $\rho = 1$. In each iteration, we use the SGD optimizer with the learning rate of $\gamma=0.01$ and clients perform $10$ local training epochs with a batch size of $B = 32$ across a total of $100$ communication rounds. The feature and prototype dimensions are set to $d_z = 512$. The hyperparameters $\tau$, $\lambda_1$, $\lambda_2$, $\lambda_3$, $\eta$ and $K$ are set to 0.07, 1.0, 10.0, 1.0, 1.0 and 30, respectively.

\subsection{Experiment Results}

{\bf Overall Performance.} We first evaluate FedDBP against existing HFL methods in heterogeneous settings. Table \ref{tab:main} shows the average accuracy, where FedDBP reaches the best performance on all four datasets, improving over FedProto by $6.95$ percentage points. These results demonstrate the generalization and stability of FedDBP. In addition, FedDBP outperforms existing federated prototype learning methods due to its dual-branch design, which balances feature fidelity and discriminability. Compared with other HFL methods such as FedKD, the superiority of FedDBP highlights the advantage of feature-level alignment over logit-level distillation, while personalized global prototypes address client-specific needs.

\begin{figure}[h]   % 不要加 * ，这样就是单栏浮动体
\setlength{\abovecaptionskip}{0pt}
\setlength{\belowcaptionskip}{0pt}
  \centering
  \includegraphics[width=\linewidth]{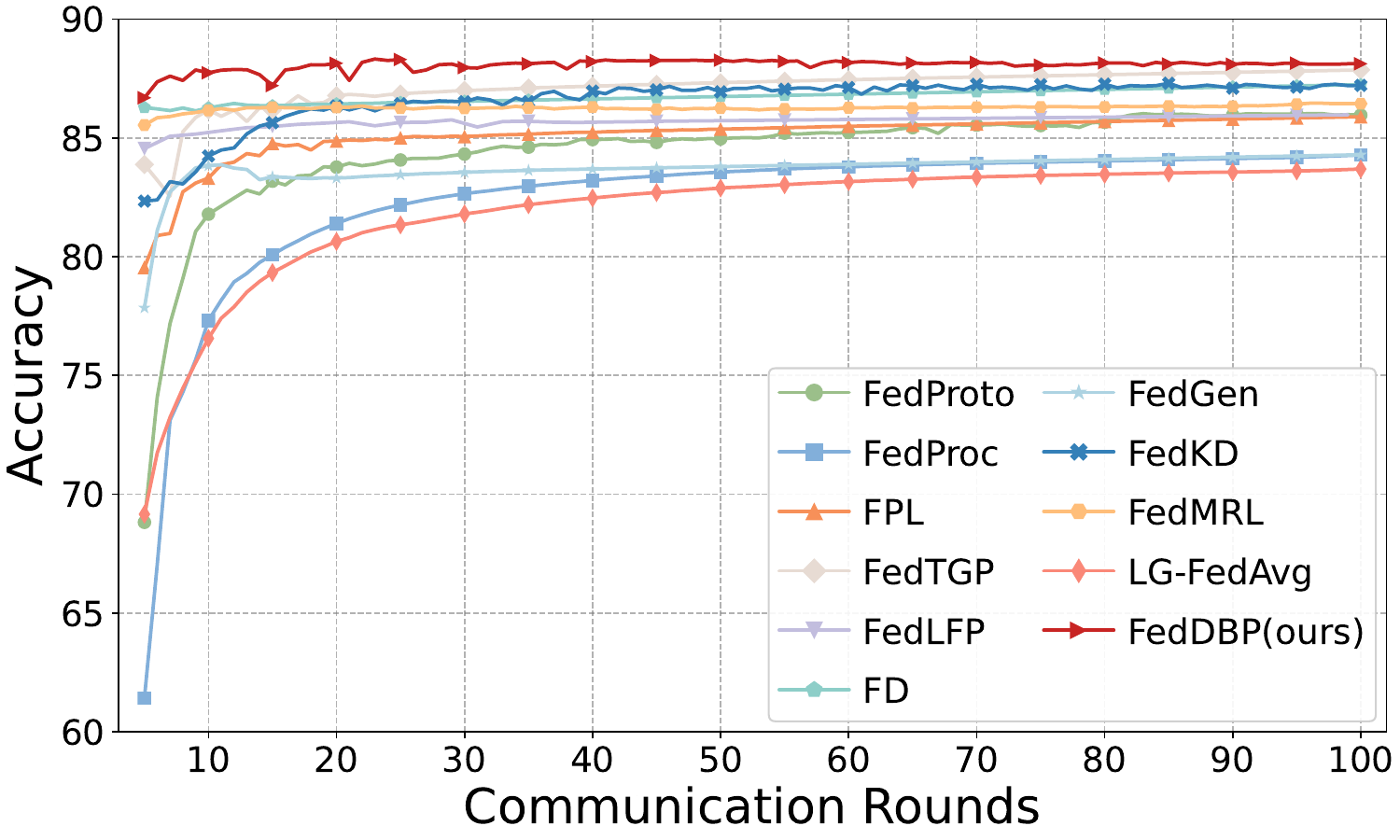}
  \caption{Visualization of training curves of the average accuracy(\%) of FedDBP and other HFL methods on the CIFAR-10 dataset.}
  \label{fig:singlecol-pdf}
\end{figure}

\begin{figure}[h]   % 不要加 * ，这样就是单栏浮动体
\setlength{\abovecaptionskip}{-5mm}
\setlength{\belowcaptionskip}{-5mm}
  \centering
  \includegraphics[width=\linewidth]{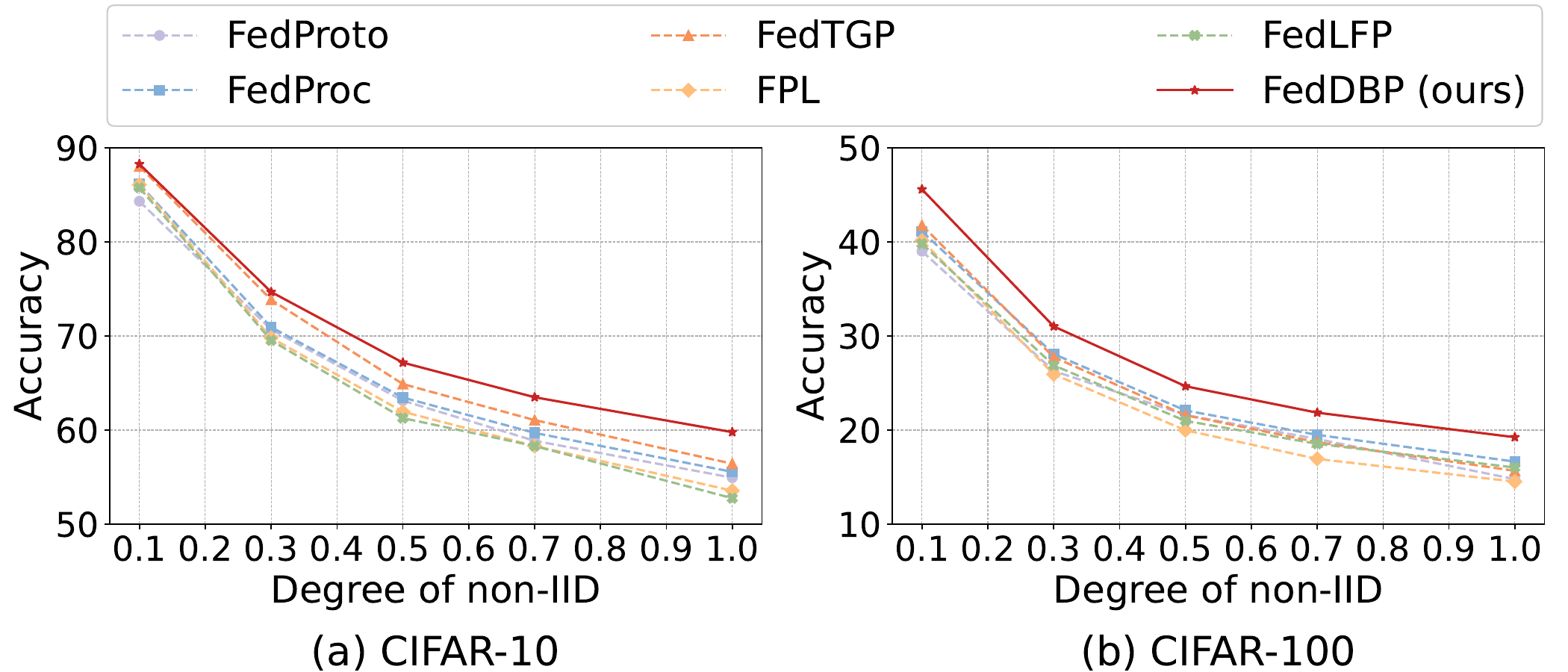}
  \caption{Visualization of accuracy(\%) of our method and others on different degrees of non-IID, which is indicated by the $\alpha$ parameter of the Dirichlet distribution.}
  \label{fig:noniid-pdf}
\end{figure}

\begin{table}[h]
\setlength{\abovecaptionskip}{0pt}
\setlength{\belowcaptionskip}{0pt}
\centering
% 调大间距
\setlength\tabcolsep{17pt}   % 默认大约 6pt，可以适当加大
\renewcommand{\arraystretch}{1.25} % 行距加大一点

% —— 仅对本表加粗线条（横线+竖线） ——
\begingroup
\setlength{\arrayrulewidth}{0.4pt}   % 粗细可调：0.6–1.0pt
\arrayrulecolor{black}
\caption{Comparison with federated prototype learning methods on CIFAR-100 dataset with different number of training epochs(\textit{E}).}
\label{table_epoch}
\begin{tabular}{rclclcl}
\Xhline{1.2pt} 
\rowcolor[HTML]{D0CECE}
\multicolumn{1}{c|}{\cellcolor[HTML]{D0CECE}} &
\multicolumn{2}{c}{\cellcolor[HTML]{D0CECE}\textit{E=1}} &
\multicolumn{2}{c}{\cellcolor[HTML]{D0CECE}\textit{E=10}} &
\multicolumn{2}{c}{\cellcolor[HTML]{D0CECE}\textit{E=20}} \\
\hline
\addlinespace[2pt]
\hline   % ← 保留你的这一句

\multicolumn{1}{r|}{FedProto} &
\multicolumn{2}{c}{\num{34.67}} &
\multicolumn{2}{c}{\num{39.01}} &
\multicolumn{2}{c}{\num{38.65}} \\

\rowcolor[HTML]{E7E6E6}
\multicolumn{1}{r|}{\cellcolor[HTML]{E7E6E6}FedProc} &
\multicolumn{2}{c}{\cellcolor[HTML]{E7E6E6}\num{41.22}} &
\multicolumn{2}{c}{\cellcolor[HTML]{E7E6E6}\num{41.12}} &
\multicolumn{2}{c}{\cellcolor[HTML]{E7E6E6}\num{41.83}} \\

\multicolumn{1}{r|}{FPL} &
\multicolumn{2}{c}{\num{38.83}} &
\multicolumn{2}{c}{\num{40.15}} &
\multicolumn{2}{c}{\num{39.53}} \\

\rowcolor[HTML]{E7E6E6}
\multicolumn{1}{r|}{\cellcolor[HTML]{E7E6E6}FedTGP} &
\multicolumn{2}{c}{\cellcolor[HTML]{E7E6E6}\num{38.21}} &
\multicolumn{2}{c}{\cellcolor[HTML]{E7E6E6}\num{41.77}} &
\multicolumn{2}{c}{\cellcolor[HTML]{E7E6E6}\num{43.03}} \\

\hline
\rowcolor[HTML]{DDEBF7}
\multicolumn{1}{r|}{\cellcolor[HTML]{DDEBF7}\bfseries FedDBP} &
\multicolumn{2}{c}{\cellcolor[HTML]{DDEBF7}\best{\num{42.59}}} &
\multicolumn{2}{c}{\cellcolor[HTML]{DDEBF7}\best{\num{45.59}}} &
\multicolumn{2}{c}{\cellcolor[HTML]{DDEBF7}\best{\num{45.26}}} \\
\Xhline{1.2pt} 
\end{tabular}
\endgroup
\end{table}

{\bf Analysis of Training Process.} Fig. \ref{fig:singlecol-pdf} presents the accuracy of the test in communication rounds $5$-$100$. From round $5$ onward, FedDBP consistently achieves higher accuracy than comparative methods, demonstrating its effectiveness. In terms of convergence, FedDBP also performs strongly. Despite occasional fluctuations in some rounds, the accuracy curve stabilizes after about $25$ rounds.

{\bf Impact of Different Degrees of Non-IID.} We evaluate FedDBP and existing federated prototype learning methods under different degrees of non-IID conditions on CIFAR-10 and CIFAR-100. As shown in Fig. \ref{fig:noniid-pdf}, FedDBP consistently outperforms other  baselines across all settings, which demonstrates the effectiveness of our method.

{\bf Impact of Number of Local Training Epochs.} As shown in Table \ref{table_epoch}, we evaluate FedDBP and existing Federated Prototype learning methods on CIFAR-100 under different numbers of local epochs. With $E=1$, which simulates frequent communication, FedDBP outperforms all methods, indicating faster convergence. With $E=20$, which simulates extensive local training, FedProc and FedTGP reach their best results, suggesting both can mitigate client drift. Nevertheless, their accuracy remains below FedDBP. Although the accuracy of FedDBP drops slightly, it remains the best, indicating a higher performance.

{\bf Ablation Study.} We evaluate the FedDBP of its individual components. Specifically, we remove the shared branch (w/o share), the decision branch (w/o deci.), hard negative mining (w/o hard), and personalized global prototypes (w/o pers.). Table  \ref{table_ablation} presents the results. The w/o share and w/o deci. degrade performance, validating the design of dual-branch feature projector. The accuracy of w/o hard also decreases. The effect is pronounced on multi-class datasets, where the w/o hard variant suffers a $10–20$ percent relative decline compared with the entire model, underscoring the necessity of hard negative mining for multi-class problems. Finally, w/o pers. also reduces performance, showing that personalized global prototypes address client-specific needs and yield improved performance.

\begin{table}[h]
\setlength{\abovecaptionskip}{0pt}
\setlength{\belowcaptionskip}{0pt}
\centering
\renewcommand{\arraystretch}{1.25} % 行距更宽
\setlength{\tabcolsep}{4pt}        % 列距更宽
\caption{Effectiveness of key components on four datasets.}
\begin{tabular}{l|clclclcl}
\Xhline{1.2pt} 
\rowcolor[HTML]{D0CECE} 
 & \multicolumn{2}{c}{\cellcolor[HTML]{D0CECE}\textit{CIFAR10}} &
   \multicolumn{2}{c}{\cellcolor[HTML]{D0CECE}\textit{CIFAR100}} &
   \multicolumn{2}{c}{\cellcolor[HTML]{D0CECE}\textit{Flowers102}} &
   \multicolumn{2}{c}{\cellcolor[HTML]{D0CECE}\textit{Tiny-ImageNet}} \\
\hline
\addlinespace[2pt]
\hline   % ← 保留你的这一句

w/o share & \multicolumn{2}{c}{\num{87.68}} &
            \multicolumn{2}{c}{\num{40.65}} &
            \multicolumn{2}{c}{\num{42.56}} &
            \multicolumn{2}{c}{\num{19.11}} \\
            
\rowcolor[HTML]{E7E6E6}
    {\cellcolor[HTML]{E7E6E6}w/o deci.}  &
    \multicolumn{2}{c}{\cellcolor[HTML]{E7E6E6}{\num{85.95}}} &
    \multicolumn{2}{c}{\cellcolor[HTML]{E7E6E6}{\num{38.42}}} &
    \multicolumn{2}{c}{\cellcolor[HTML]{E7E6E6}{\num{44.3}}} &
    \multicolumn{2}{c}{\cellcolor[HTML]{E7E6E6}{\num{20.81}}} \\

w/o hard & \multicolumn{2}{c}{\num{86.94}} &
             \multicolumn{2}{c}{\num{42.76}} &
             \multicolumn{2}{c}{\num{48.16}} &
             \multicolumn{2}{c}{\num{23.22}} \\

\rowcolor[HTML]{E7E6E6}
    {\cellcolor[HTML]{E7E6E6}w/o pers.} & 
    \multicolumn{2}{c}{\cellcolor[HTML]{E7E6E6}{\num{88.02}}} &
    \multicolumn{2}{c}{\cellcolor[HTML]{E7E6E6}{\num{44.68}}} &
    \multicolumn{2}{c}{\cellcolor[HTML]{E7E6E6}{\num{52.6}}} &
    \multicolumn{2}{c}{\cellcolor[HTML]{E7E6E6}{\num{26.94}}} \\
\hline

\rowcolor[HTML]{DDEBF7} 
\bfseries FedDBP & \multicolumn{2}{c}{\cellcolor[HTML]{DDEBF7}\best{\num{88.27}}} &
       \multicolumn{2}{c}{\cellcolor[HTML]{DDEBF7}\best{\num{45.59}}} &
       \multicolumn{2}{c}{\cellcolor[HTML]{DDEBF7}\best{\num{52.99}}} &
       \multicolumn{2}{c}{\cellcolor[HTML]{DDEBF7}\best{\num{27.14}}} \\
\Xhline{1.2pt} 
\end{tabular}
\label{table_ablation}
\end{table}

\section{CONCLUSION}

This paper proposes a novel FPL method FedDBP, including a dual-branch feature projector on the client side and a personalized global prototype fusion on the server. These processes balance the fidelity and discriminability of local feature, and achieve adaptive personalization of global prototype. 
Extensive experiments show that FedDBP significantly excels over existing state-of-the-art federated prototype learning methods under different heterogeneous settings.

\clearpage
\small

% \section{Acknowledgements} 
% This work is supported by the National Natural Science Foundation of China (Grant No. 62476095), Guangdong Basic and Applied Basic Research Foundation (Grant No. 2025A1515011525)
% References should be produced using the bibtex program from suitable
% BiBTeX files (here: strings, refs, manuals). The IEEEbib.bst bibliography
% style file from IEEE produces unsorted bibliography list.
% -------------------------------------------------------------------------
\bibliographystyle{IEEEbib}
\bibliography{refs}

\end{document}